\documentclass{article}
 
\usepackage[utf8]{inputenc}
\usepackage[T1]{fontenc}
\usepackage{amsmath,amssymb,amsfonts}
\usepackage{graphicx}
\usepackage{booktabs}
\usepackage{hyperref}
\usepackage{xcolor}
\usepackage{subcaption}
\usepackage[margin=1in]{geometry}
\usepackage{natbib}
\usepackage{algorithm}
\usepackage{algorithmic}

\definecolor{mlapink}{HTML}{E91E63}
\definecolor{mlablue}{HTML}{2196F3}
\definecolor{mlapurple}{HTML}{9C27B0}

\title{\textbf{Through the Bottleneck:} \\
How Multi-head Latent Attention Separates Content \\
from Position in Language Models}

\author{
  \textbf{Dhruvil S}\thanks{Equal contribution. \\ \indent Code and data available at \url{https://github.com/Dhruvil-sr24/Small-Language-Model-From-Scratch/tree/main/interp}}\textsuperscript{\, ,1}
  \quad \quad \quad
  \textbf{Fenil Sojitra}\footnotemark[1]\textsuperscript{\, ,2} \\[0.2cm]
  \textbf{Ravirajsinh Chauhan}\textsuperscript{2} \\[0.2cm]
  \textsuperscript{1}Indian Institute of Technology Madras \\
  \textsuperscript{2}P P Savani University \\
  \texttt{23f2004889@ds.study.iitm.ac.in} \quad \texttt{22se02it061@ppsu.ac.in} \\
  \texttt{raviraj.chauhan@ppsu.ac.in}
}
\date{}

\begin{document}

\maketitle

\begin{abstract}
Multi-head Latent Attention (MLA), introduced in DeepSeek-V2, compresses key-value pairs through a shared low-rank bottleneck ($c_\text{KV}$), achieving 81\% KV-cache reduction during inference. Despite its adoption in production models with hundreds of billions of parameters, no prior work has studied \emph{what information this bottleneck preserves or discards}, nor how it reshapes the internal circuits of a transformer. We present the first comprehensive mechanistic interpretability study of MLA, training a 114M-parameter MLA transformer (pretrained on a web/code/math mixture, fine-tuned on TinyStories) and analyzing its internal representations through four complementary experiments: singular value decomposition of compression matrices, attention head taxonomy, linear probing, and a $c_\text{KV}$ disruption-attribution analysis. Our key findings are: (1)~the $c_\text{KV}$ bottleneck learns a pure \textbf{content representation}, preserving entity identity (98\% retention) while discarding positional information (near chance accuracy), validating MLA's architectural separation of content from position via RoPE; (2)~induction heads \textbf{co-locate at a single layer} (Layer~12), unlike their distributed formation in standard MHA, suggesting the shared bottleneck constrains circuit topology; (3)~a single ``semantic hub'' layer (Layer~15) simultaneously exhibits the highest SVD effective rank and the strongest disruption-attribution score; and (4)~the bottleneck is globally \textbf{over-provisioned}, using only 46\% of its 128-dimensional capacity on average, with concrete implications for heterogeneous rank allocation. These findings, from a single 114M-parameter model trained on a narrow-domain corpus, suggest that MLA does not merely compress attention passively; it appears to reshape how the model organizes content, position, and circuit structure. We view this as an initial data point rather than a settled characterization of MLA at scale, and we detail the scope limitations in Section~\ref{sec:discussion}.
\end{abstract}

\section{Introduction}
\label{sec:intro}

Multi-head Latent Attention (MLA) \citep{deepseekv2} has rapidly become a key architectural innovation in efficient large language models. By compressing key-value (KV) representations through a shared low-rank bottleneck before projecting them into per-head keys and values, MLA achieves dramatic reductions in KV-cache memory, up to 81\% compared to standard Multi-Head Attention (MHA), without sacrificing generation quality. This design has been adopted in production-scale models including DeepSeek-V2 (236B parameters) and DeepSeek-V3 (671B parameters).

Despite the practical success of MLA and the growing field of mechanistic interpretability \citep{elhage2021mathematical, olsson2022induction, conmy2023automated, neel2022comprehensive}, \textbf{no prior work has studied what happens inside the MLA bottleneck}. Specifically:

\begin{itemize}
    \item What linguistic information does the model choose to \emph{preserve} in the compressed $c_\text{KV}$ representation?
    \item What information does it \emph{discard}, and why?
    \item How does the shared bottleneck affect the formation of known circuits (e.g., induction heads)?
    \item Which layers' bottlenecks are \emph{causally} important for the model's predictions?
\end{itemize}

We address these questions through a comprehensive mechanistic study of a 114M-parameter MLA transformer, pretrained on a mixture of web text, code, and mathematical text and then fine-tuned on TinyStories \citep{eldan2023tinystories} (see Section~\ref{sec:setup} for the full data mixture).Our four experiments (SVD analysis, attention head taxonomy, linear probing, and $c_\text{KV}$ disruption attribution) converge on a coherent picture of how MLA reshapes transformer information processing.

\paragraph{Contributions.}
\begin{enumerate}
    \item We show that $c_\text{KV}$ learns a \textbf{content-only representation}: entity information is preserved with 98\% fidelity, while positional information is discarded (near-chance probe accuracy). This validates MLA's design of separating content ($c_\text{KV}$) from position (RoPE $k_\text{rope}$).
    \item We discover that induction heads \textbf{co-locate at a single layer} in MLA (all 5 detected induction heads at Layer~12), in contrast to their distributed formation in standard MHA \citep{olsson2022induction}.
    \item We identify a \textbf{semantic hub} at Layer~15 that simultaneously has the highest SVD effective rank (88/128) and the highest disruption-attribution score among all layers.
    \item We demonstrate that the bottleneck is \textbf{globally over-provisioned}: only 46\% of the 128-dim capacity is used on average. We propose heterogeneous rank allocation as a concrete architectural improvement.
\end{enumerate}

\section{Background}
\label{sec:background}

\subsection{Multi-head Latent Attention}

Standard MHA computes queries, keys, and values independently for each head from the residual stream $\mathbf{x} \in \mathbb{R}^{d}$:
\begin{equation}
    \mathbf{q}_h = W_Q^h \mathbf{x}, \quad
    \mathbf{k}_h = W_K^h \mathbf{x}, \quad
    \mathbf{v}_h = W_V^h \mathbf{x}
\end{equation}

This requires caching $\mathbf{k}_h$ and $\mathbf{v}_h$ for all heads during autoregressive generation, costing $O(n_\text{heads} \cdot d_\text{head})$ memory per token per layer.

MLA introduces a \textbf{shared KV compression bottleneck}:
\begin{align}
    \mathbf{c}_\text{KV} &= \text{RMSNorm}(W_\text{DKV} \cdot \mathbf{x}) \in \mathbb{R}^{r_\text{KV}} \label{eq:ckv} \\
    \mathbf{k}_h^\text{nope} &= W_{UK}^h \cdot \mathbf{c}_\text{KV}, \quad
    \mathbf{v}_h = W_{UV}^h \cdot \mathbf{c}_\text{KV} \\
    \mathbf{k}_h^\text{rope} &= \text{RoPE}(W_{KR}^h \cdot \mathbf{x}) \label{eq:krope} \\
    \mathbf{k}_h &= [\mathbf{k}_h^\text{nope}; \mathbf{k}_h^\text{rope}]
\end{align}

During inference, only $\mathbf{c}_\text{KV} \in \mathbb{R}^{r_\text{KV}}$ needs to be cached (rather than all per-head K/V), achieving compression ratio $r_\text{KV} / (n_\text{heads} \cdot d_\text{head})$.

Critically, Eq.~\ref{eq:ckv} and \ref{eq:krope} reveal an architectural separation: \textbf{content information} flows through $\mathbf{c}_\text{KV}$ (shared across heads), while \textbf{positional information} flows through the separate RoPE pathway $\mathbf{k}_h^\text{rope}$ (directly from $\mathbf{x}$, bypassing the bottleneck). Whether the model actually learns to exploit this separation is an empirical question we answer in Section~\ref{sec:probing}.

\subsection{Mechanistic Interpretability}

We build on established mechanistic interpretability techniques:
\begin{itemize}
    \item \textbf{Linear probing} \citep{belinkov2017neural, alain2016understanding}: Training a linear classifier on intermediate representations to test what information they encode: if a linear probe achieves high accuracy, the information is linearly accessible in the representation.
    \item \textbf{Activation patching} \citep{meng2022locating, wang2023interpretability}: the classic version of this technique corrupts inputs and then surgically restores clean activations at specific (layer, position) pairs, measuring how much output is recovered, establishing genuinely causal (not merely correlational) importance. In Section~\ref{sec:patching}, we use a lighter-weight variant, $c_\text{KV}$ \textbf{disruption attribution}, which measures how much corruption \emph{changes} $c_\text{KV}$ at each (layer, position) and reweights this by the resulting drop in output probability, without performing the restoration step. This is cheaper to compute but establishes correlational, not strictly causal, importance; we flag this distinction explicitly where it matters (Section~\ref{sec:limitations}).
    \item \textbf{Induction head analysis} \citep{olsson2022induction}: Identifying attention heads that implement the ``induction'' circuit (copying tokens that previously followed a matching prefix), a fundamental in-context learning mechanism.
\end{itemize}

\section{Experimental Setup}
\label{sec:setup}

\subsection{Model Architecture}

Our model is a 24-layer MLA transformer with the following configuration:

\begin{table}[h]
\centering
\caption{Model architecture details.}
\begin{tabular}{lcc}
\toprule
\textbf{Parameter} & \textbf{Value} & \textbf{Notes} \\
\midrule
Layers & 24 & All transformer blocks \\
Hidden dim ($d_\text{model}$) & 512 & Residual stream width \\
Attention heads ($n_h$) & 8 & \\
KV heads ($n_\text{kv}$) & 8 & Grouped Query Attention \\
Head dim ($d_\text{head}$) & 64 & $d_\text{nope} + d_\text{rope}$ \\
KV bottleneck ($r_\text{KV}$) & 128 & \textbf{Key: 4$\times$ compression} \\
Q bottleneck ($r_Q$) & 256 & 2$\times$ compression \\
QK nope dim & 48 & Content dimension \\
QK rope dim & 16 & Position dimension \\
Total parameters & 114.1M & \\
KV-cache reduction & 81\% & vs.\ standard MHA \\
\bottomrule
\end{tabular}
\label{tab:architecture}
\end{table}

\subsection{Training}

The model was pretrained for 16,500 steps on a mixed corpus: 60\% FineWeb-Edu \citep{penedo2024fineweb} (general web text, for foundational language and grammar), 25\% The Stack \citep{kocetkov2022stack} (source code, for coding logic and structured reasoning), and 15\% OpenWebMath \citep{paster2023openwebmath} (mathematical web text, for introductory-level mathematical language). It was then fine-tuned for 7,200 steps on TinyStories \citep{eldan2023tinystories}, achieving a perplexity of 4.3 on the TinyStories validation set. All interpretability analyses in this paper are run on the \emph{fine-tuned} checkpoint, using TinyStories-distribution passages as the analysis input unless otherwise stated, so while the model’s representations were shaped by a broader and more heterogeneous pretraining mixture (web text, code, and math) than TinyStories alone, the probing, attention-taxonomy, and disruption-attribution results reflect how the model behaves once specialized to the simpler TinyStories domain. We use a BPE tokenizer with vocabulary size 32,768.

\subsection{Experiments Overview}

We conduct four experiments, each probing a different aspect of the MLA bottleneck:

\begin{enumerate}
    \item \textbf{SVD Analysis} (Section~\ref{sec:svd}): Decompose compression matrices $W_\text{DKV}$ and $W_\text{DQ}$ to measure effective dimensionality.
    \item \textbf{Attention Patterns} (Section~\ref{sec:attention}): Classify all $24 \times 8 = 192$ heads by function (previous-token, induction, BOS-sink, local window).
    \item \textbf{Linear Probing} (Section~\ref{sec:probing}): Compare what linguistic features are linearly decodable from $\mathbf{c}_\text{KV}$ (128-dim) vs.\ the residual stream (512-dim).
    \item \textbf{$c_\text{KV}$ Disruption Attribution} (Section~\ref{sec:patching}): Measure how strongly corruption-induced changes to $\mathbf{c}_\text{KV}$ at each layer correlate with degraded predictions, as a lightweight proxy for causal importance.
\end{enumerate}

All experiments are inference-only and run on a single NVIDIA T4 GPU.

\section{Results}

\subsection{SVD Analysis: The Bottleneck is Over-Provisioned}
\label{sec:svd}

We compute the SVD of $W_\text{DKV} \in \mathbb{R}^{128 \times 512}$ at each of the 24 layers and measure the \emph{effective rank}: the number of singular values needed to capture 99\% of the total energy ($\sum \sigma_i^2$).

\begin{table}[h]
\centering
\caption{Effective rank of KV and Q compression matrices.}
\begin{tabular}{lccc}
\toprule
\textbf{Matrix} & \textbf{Mean Rank} & \textbf{Min (Layer)} & \textbf{Max (Layer)} \\
\midrule
$W_\text{DKV}$ (512 $\to$ 128) & 58.8 / 128 (46\%) & 44 (L6) & 88 (L15) \\
$W_\text{DQ}$ (512 $\to$ 256) & 99.0 / 256 (39\%) & 41 (L15) & 130 (L21) \\
\bottomrule
\end{tabular}
\label{tab:svd}
\end{table}

\paragraph{Key finding:} On average, the model uses only \textbf{46\% of its 128-dim KV bottleneck capacity}. At the 90\% energy threshold, only 17 dimensions suffice on average—suggesting significant over-provisioning.

\paragraph{Layer heterogeneity:} Three layers exhibit markedly higher effective rank: Layer~7 (85), Layer~11 (79), and Layer~15 (88). These ``rank spikes'' correspond to architectural phase transitions, as we show in subsequent experiments.

\paragraph{Anti-correlation of KV and Q ranks:} When $W_\text{DKV}$ uses more dimensions (e.g., Layer~15: rank 88), $W_\text{DQ}$ uses fewer (Layer~15: rank 41), and vice versa. This suggests the model learns an \textbf{asymmetric capacity allocation}: when more information is needed about what positions attend \emph{to} (KV), less is needed about what positions attend \emph{from} (Q).

\begin{figure}[h]
\centering
\includegraphics[width=0.85\textwidth]{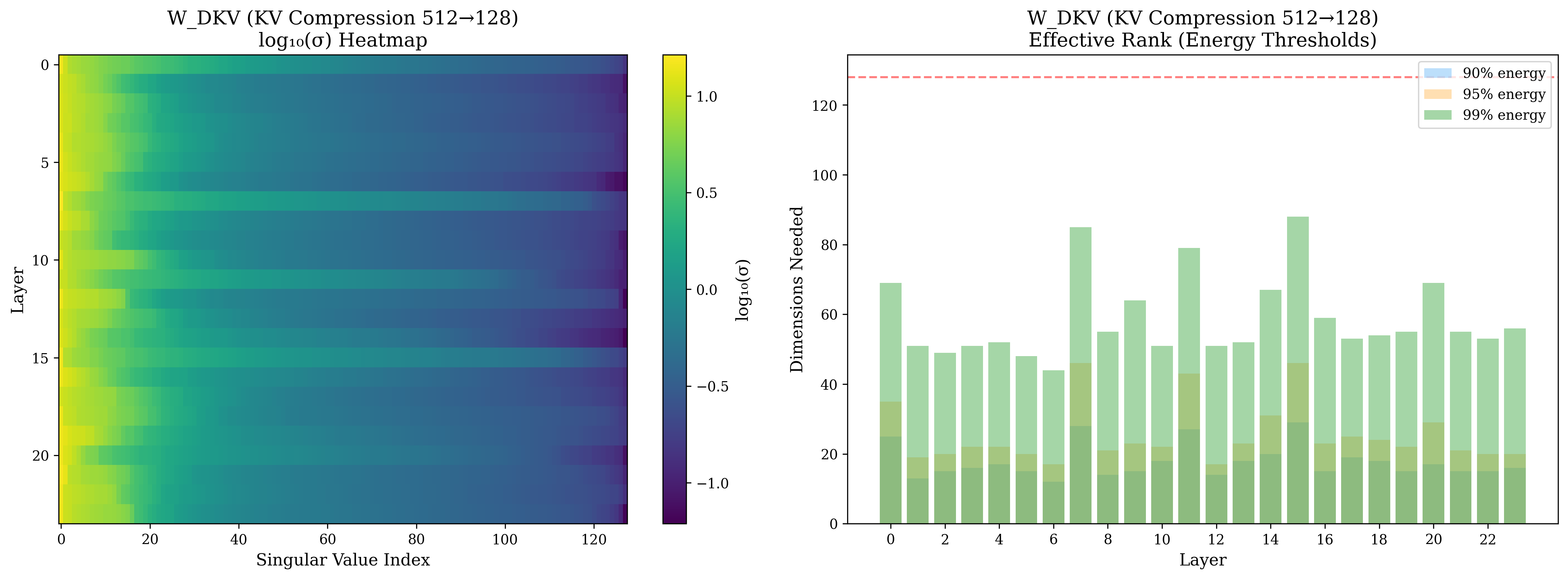}
\caption{Left: log singular-value magnitude of $W_\text{DKV}$ by layer, showing rapid decay after roughly the first 20--30 components at every layer. Right: effective rank at 90/95/99\% energy thresholds per layer, all well below the full 128-dim capacity (dashed line).}
\label{fig:svd-spectrum}
\end{figure}

\begin{figure}[h]
\centering
\includegraphics[width=0.85\textwidth]{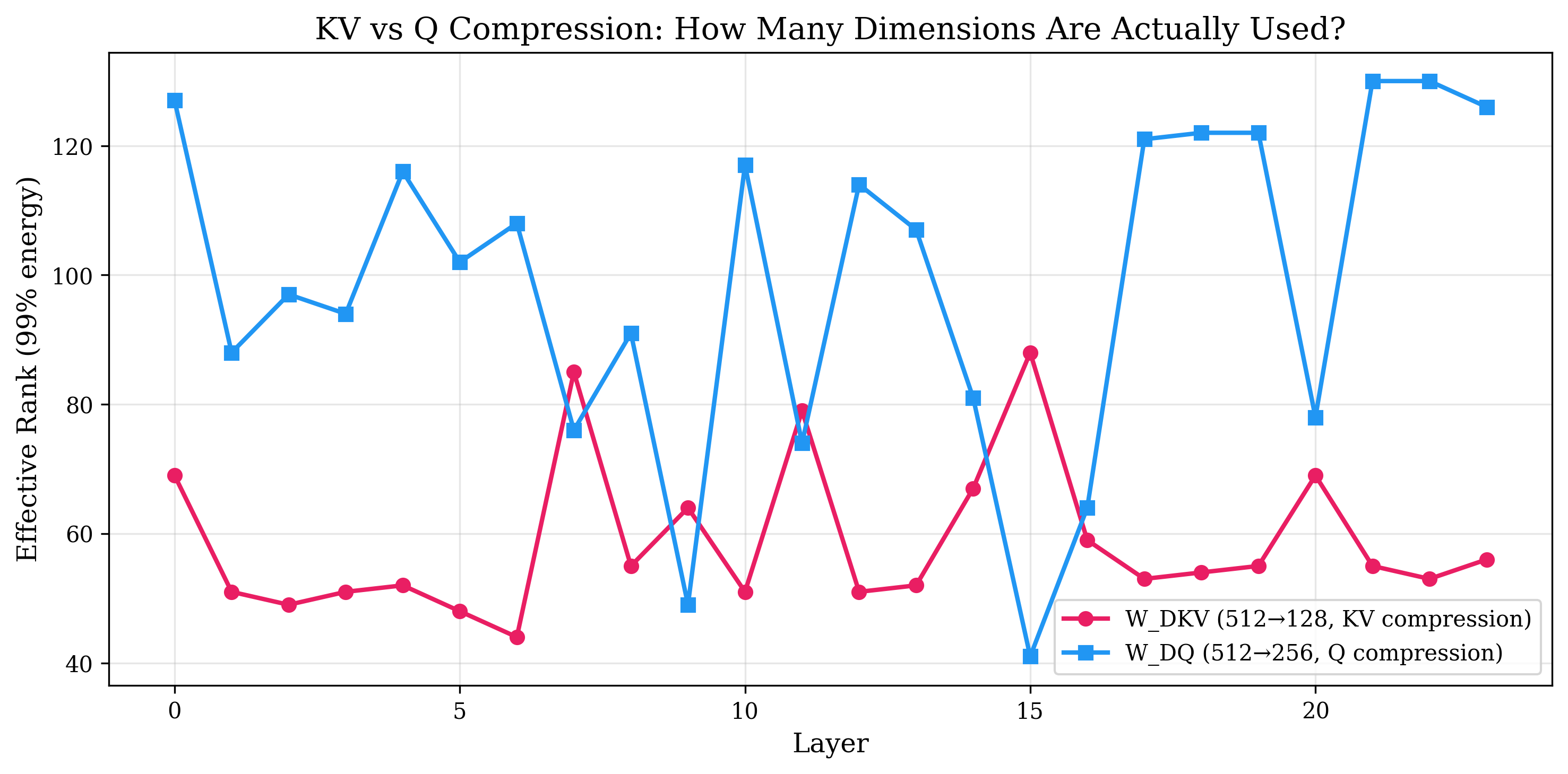}
\caption{Effective rank (99\% energy) of $W_\text{DKV}$ (KV compression) vs.\ $W_\text{DQ}$ (Q compression) across layers, illustrating the anti-correlation pattern.}
\label{fig:svd-anticorr}
\end{figure}

\subsection{Attention Patterns: Induction Heads Co-locate}
\label{sec:attention}

We classify all 192 attention heads using four metrics computed on 50 TinyStories passages (sequence length 64): (1)~\emph{previous-token score}: fraction of attention on position $t-1$; (2)~\emph{induction score}: attention on positions following a matching prefix; (3)~\emph{BOS score}: attention on the beginning-of-sequence token; (4)~\emph{local window score}: attention within a 5-token window.

\begin{table}[h]
\centering
\caption{Top induction and previous-token heads.}
\begin{tabular}{lcc}
\toprule
\textbf{Head} & \textbf{Score} & \textbf{Type} \\
\midrule
L12 H6 & 0.250 & Induction \\
L12 H5 & 0.240 & Induction \\
L12 H7 & 0.204 & Induction \\
L12 H2 & 0.202 & Induction \\
L12 H3 & 0.194 & Induction \\
\midrule
L3 H3 & 0.971 & Previous-token \\
L10 H6 & 0.953 & Previous-token \\
L2 H2 & 0.948 & Previous-token \\
\bottomrule
\end{tabular}
\label{tab:heads}
\end{table}

\paragraph{Key finding:} All five detected induction heads are at \textbf{Layer~12 exclusively}. In standard MHA, \citet{olsson2022induction} found induction heads distributed across multiple layers in 2+ layer models. We hypothesize that MLA's shared $\mathbf{c}_\text{KV}$ bottleneck constrains induction heads to co-locate: since the induction circuit requires coordinated K-V computation, and both K and V are derived from the same compressed representation, it is more efficient for all induction heads to operate on the same layer's $\mathbf{c}_\text{KV}$.

Previous-token heads, which implement a simpler circuit (attend to position $t-1$), form across early layers (L1--L5) as expected. This creates a clear \textbf{previous-token $\to$ induction} pipeline, consistent with the two-step induction mechanism \citep{olsson2022induction}, but compressed into fewer layers.

\begin{figure}[h]
\centering
\includegraphics[width=0.95\textwidth]{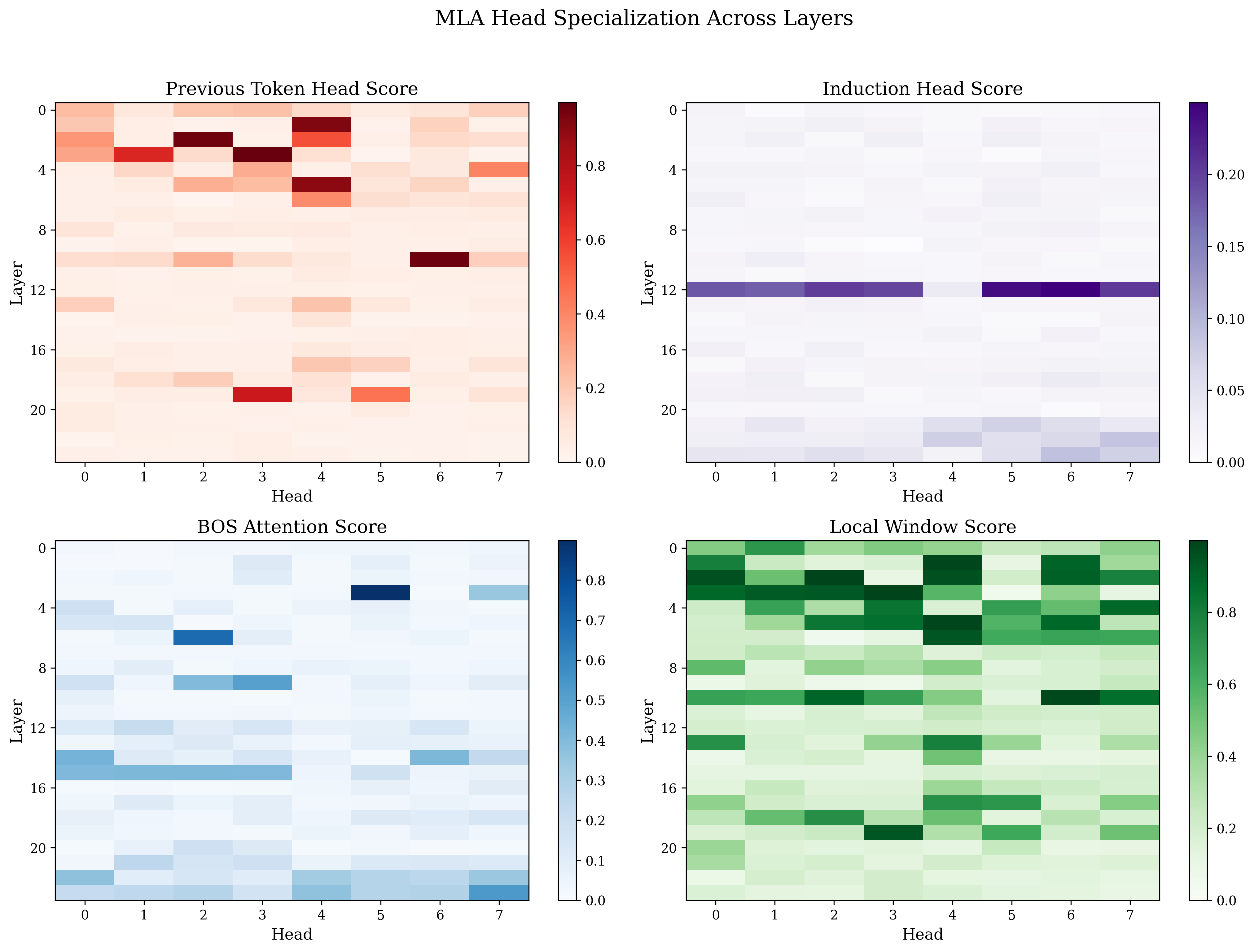}
\caption{All four head-classification scores (previous-token, induction, BOS-attention, local-window) across every layer and head. The induction-score panel shows a sharp, isolated band at Layer~12 with no comparable signal at any other layer.}
\label{fig:head-heatmaps}
\end{figure}

\begin{figure}[h]
\centering
\includegraphics[width=0.7\textwidth]{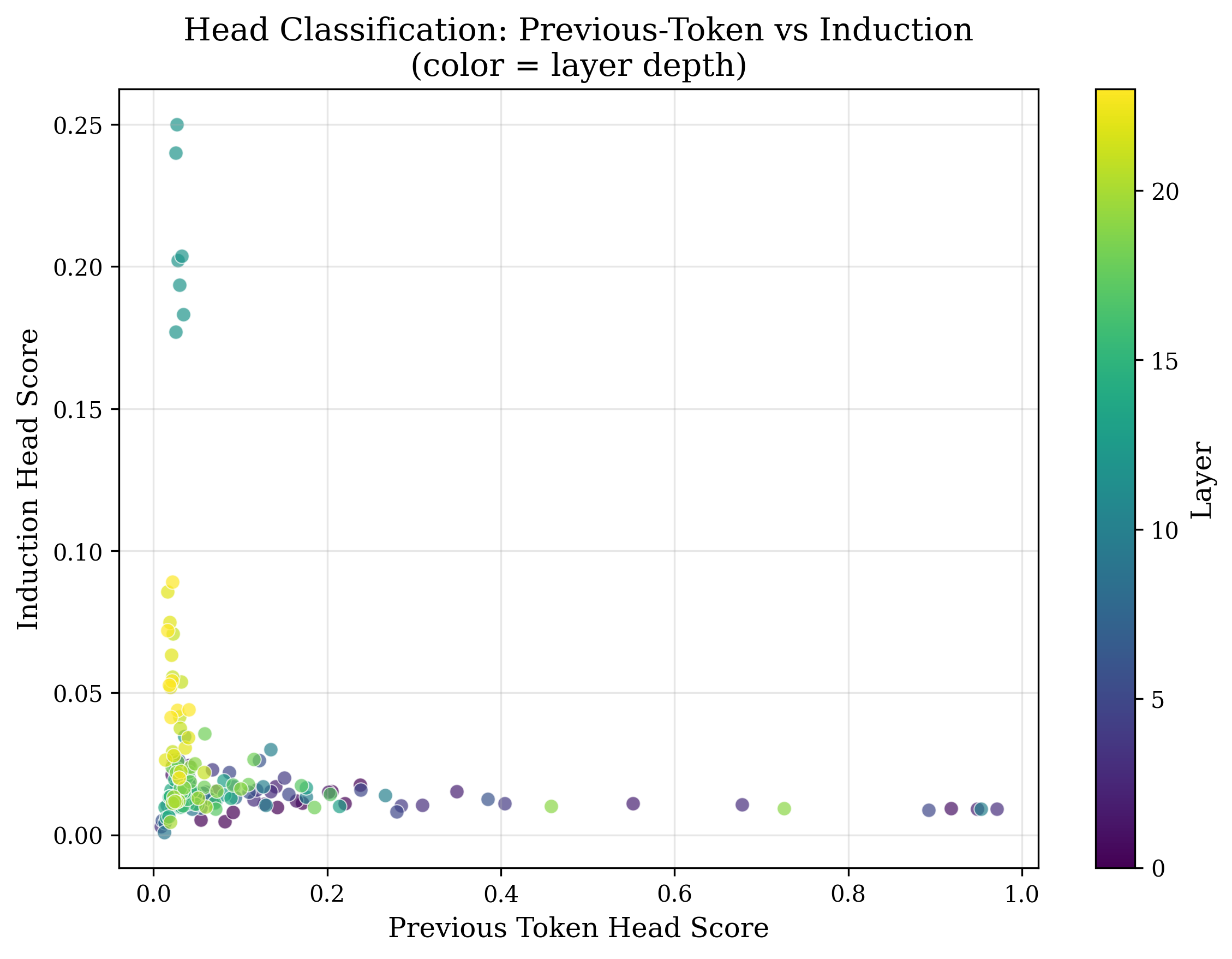}
\caption{Previous-token score vs.\ induction score for all 192 heads, colored by layer depth. The cluster of high-induction-score points in the teal range corresponds entirely to Layer~12 heads.}
\label{fig:head-scatter}
\end{figure}

\subsection{Linear Probing: Content-Position Separation}
\label{sec:probing}

We train linear probes on $\mathbf{c}_\text{KV}$ (128-dim) and the residual stream (512-dim) to classify five linguistic properties at each of 8 layers (0, 4, 8, 11, 12, 15, 20, 23). Probes are trained on 12,800 token positions (100 passages $\times$ 128 tokens) with a 80/20 train/val split.

\begin{table}[h]
\centering
\caption{Probing accuracy (validation) and bottleneck retention at Layer~12.}
\begin{tabular}{lcccc}
\toprule
\textbf{Probe} & \textbf{Classes} & \textbf{c\_KV Acc} & \textbf{Residual Acc} & \textbf{Retention} \\
\midrule
Entity (is character?) & 2 & 0.955 & 0.955 & \textbf{100.0\%} \\
POS tag & 7 & 0.891 & 0.975 & 91.3\% \\
Token identity & 32,768 & 0.734 & 0.965 & 76.0\% \\
Position (16 bins) & 16 & 0.217 & 0.372 & 58.3\%$^\dagger$ \\
Next token & 32,768 & 0.197 & 0.401 & 49.2\% \\
\bottomrule
\end{tabular}

\vspace{0.3em}
{\small $^\dagger$Position retention appears moderate, but c\_KV accuracy (21.7\%) is barely above chance (6.25\%). The retention ratio is a misleading metric here because it is computed relative to a residual-stream baseline that is itself far from ceiling; we report absolute accuracy alongside it for this reason and recommend readers weight absolute accuracy over the ratio whenever the baseline itself is weak.}
\label{tab:probing}
\end{table}

\paragraph{Key finding:} $\mathbf{c}_\text{KV}$ preserves \textbf{entity information with near-perfect fidelity} (99--100\% retention, essentially flat across all measured layers) while \textbf{discarding positional information} (c\_KV accuracy 13--22\%, near chance for 16 bins). This confirms that MLA's architectural separation(content through $\mathbf{c}_\text{KV}$, position through RoPE) is not merely structural but is actively exploited by the learned model.

\paragraph{Layer dynamics:} Entity retention is uniformly high (99.0--100.3\%) at every measured layer, from L0 through L23, with no clear depth trend; entity information appears to be preserved in $\mathbf{c}_\text{KV}$ from the earliest layers onward rather than being progressively built up. Token identity shows a more interesting non-monotonic pattern: high retention at L0 (93.5\%) falls sharply through the middle layers (35.2\% at L20) before partially recovering at the final layer (84.1\% at L23). The final-layer recovery is consistent with the output projection needing to reconstruct token identity for next-token prediction, though we note this is a two-point pattern (one dip, one recovery) rather than a smooth trend, and we have not verified it holds at intermediate layers beyond those we probed.

\begin{figure}[h]
\centering
\includegraphics[width=0.85\textwidth]{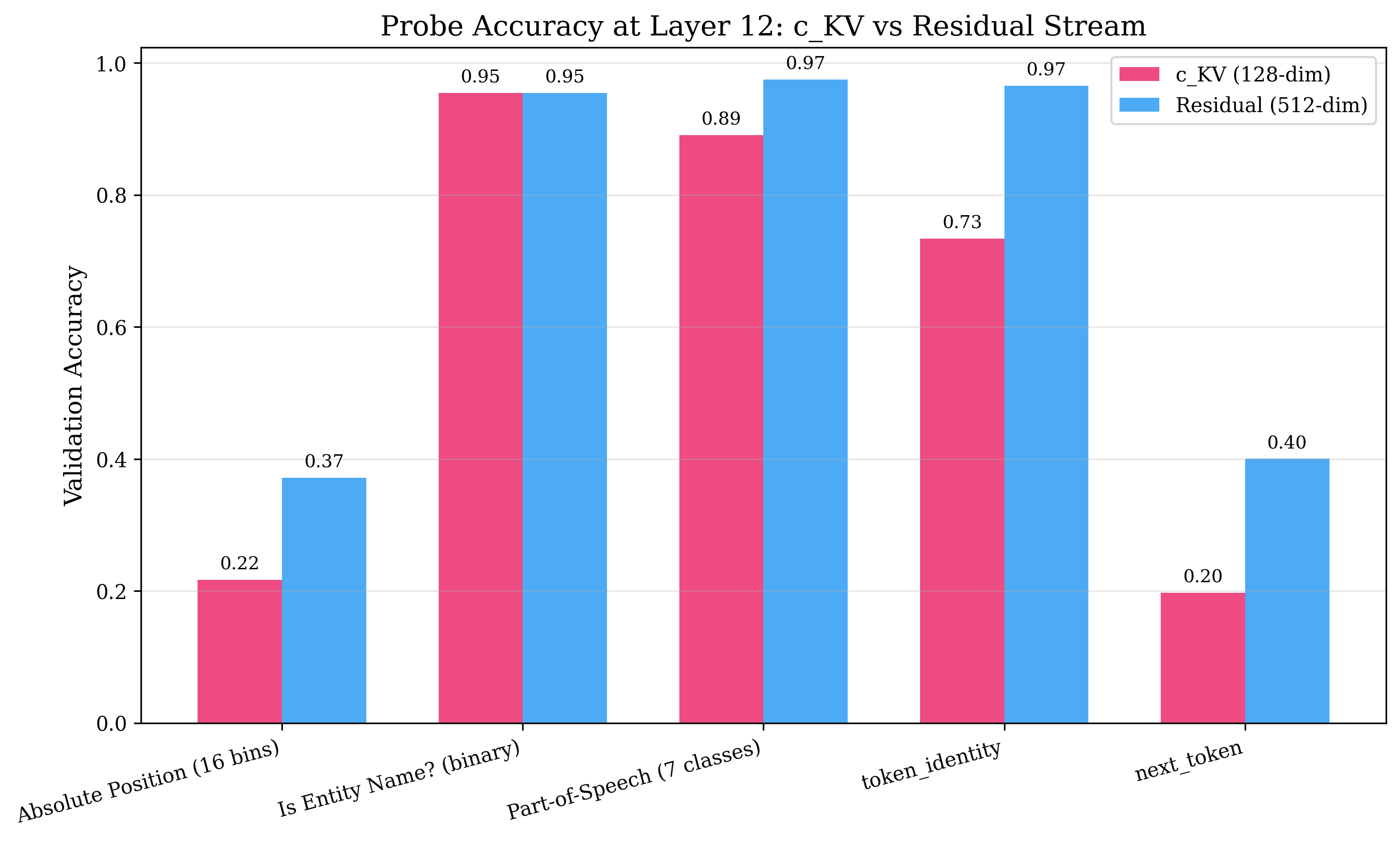}
\caption{Probe validation accuracy at Layer~12 for $\mathbf{c}_\text{KV}$ (128-dim) vs.\ the full residual stream (512-dim), across all five probed properties. Entity and POS gaps are small; position, token identity, and next-token gaps are large.}
\label{fig:probing-layer12}
\end{figure}

\begin{figure}[h]
\centering
\includegraphics[width=0.95\textwidth]{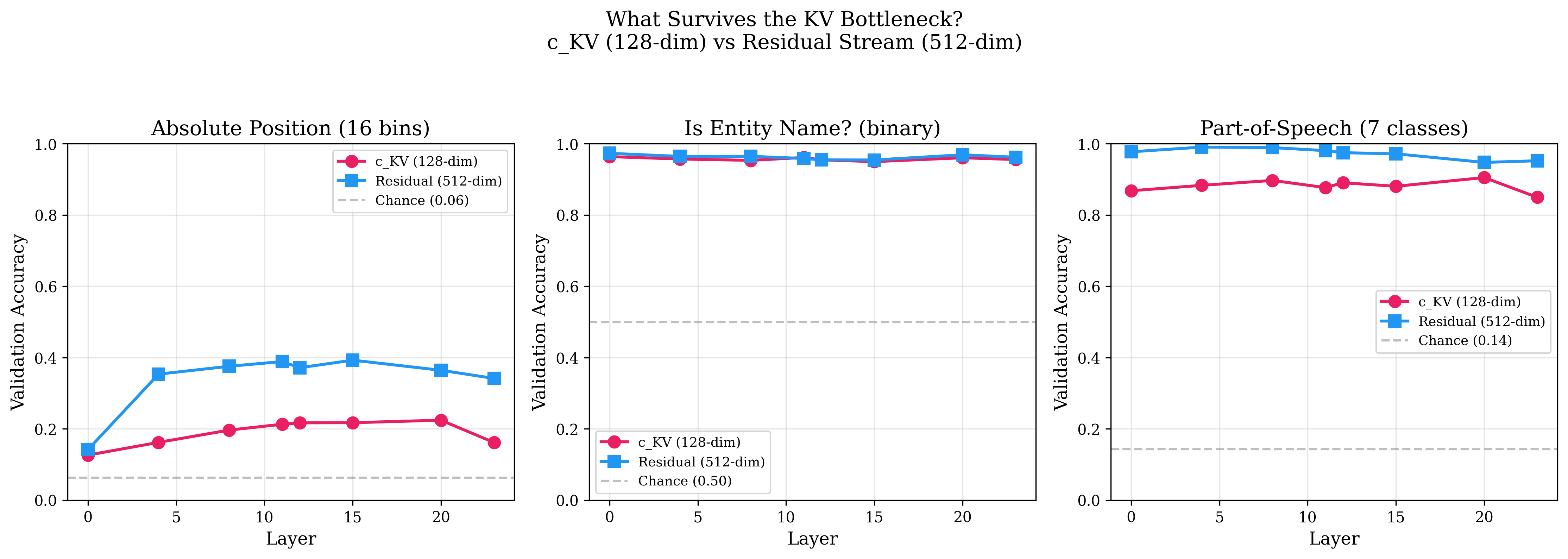}
\caption{Validation accuracy of $\mathbf{c}_\text{KV}$ vs.\ residual-stream probes across all measured layers, for absolute position, entity, and part-of-speech. Position accuracy for $\mathbf{c}_\text{KV}$ stays close to chance (dashed line) at every layer; entity and POS accuracy track the residual stream closely.}
\label{fig:probing-alllayers}
\end{figure}

\paragraph{Induction layer alignment:} Entity retention is high at every layer we measured, including Layer~12 (100.0\%), so we cannot claim it \emph{peaks specifically} at the induction-head layer; retention there is not distinguishable from other layers. What we can say is that entity information is available in $\mathbf{c}_\text{KV}$ at Layer~12, which is a necessary (but not sufficient) condition for the induction circuit to use it. We do not have direct evidence that the induction heads specifically consume entity features rather than other content encoded in $\mathbf{c}_\text{KV}$; this would require a more targeted causal experiment (e.g., ablating the entity-relevant subspace and measuring induction-head behavior) that we leave to future work.

\subsection{$c_\text{KV}$ Disruption Attribution: Layer~15 as the Semantic Hub}
\label{sec:patching}

We measure the causal relevance of $\mathbf{c}_\text{KV}$ at each layer using a disruption-attribution procedure inspired by, but distinct from, classic activation patching \citep{meng2022locating}. For each of 100 examples, we:
\begin{enumerate}
    \item Run the model cleanly, caching all $\mathbf{c}_\text{KV}$ activations.
    \item Add Gaussian noise ($\sigma=3.0$) to early token embeddings (positions 1--3), and re-run the model on the corrupted input (5 noise trials per example).
    \item For each layer, compute the cosine dissimilarity between clean and corrupted $\mathbf{c}_\text{KV}$ at every position, then scale the resulting per-(layer, position) disruption map by the trial's drop in probability on the correct next token, $\max(0, p_\text{clean} - p_\text{corrupt})$.
\end{enumerate}

Unlike classic activation patching, we do not restore clean activations into the corrupted forward pass and re-measure the output; we only measure how much corruption \emph{changes} $\mathbf{c}_\text{KV}$ and reweight that change by an aggregate, trial-level (not per-layer or per-position) probability drop. This is therefore a correlational signal (layers whose $\mathbf{c}_\text{KV}$ changes most under corruption, on examples where corruption also hurt the prediction) rather than a direct measurement of what restoring information at a specific (layer, position) would causally fix. We adopt the term \textbf{disruption attribution} rather than \emph{activation patching} to reflect this precisely, and discuss the implications in Section~\ref{sec:limitations}.

\begin{table}[h]
\centering
\caption{Disruption-attribution score of $c_\text{KV}$ by layer group (normalized, 5 trials/example). See Section~\ref{sec:patching} for how this differs from causal-patching estimates.}
\begin{tabular}{lcc}
\toprule
\textbf{Layer Group} & \textbf{Mean Attribution Score} & \textbf{Role} \\
\midrule
Layers 0--6 & 0.57 & Feature building \\
Layers 7--12 & 0.51 & Induction / transition \\
\textbf{Layers 13--19} & \textbf{0.88} & \textbf{Semantic processing} \\
Layers 20--23 & 0.73 & Output preparation \\
\midrule
\textbf{Layer 15 (peak)} & \textbf{1.00} & Semantic hub \\
Layer 7 (trough) & 0.40 & Redundant compression \\
Layer 12 (induction) & 0.50 & Moderate \\
\bottomrule
\end{tabular}
\label{tab:patching}
\end{table}

\begin{figure}[h]
\centering
\includegraphics[width=0.85\textwidth]{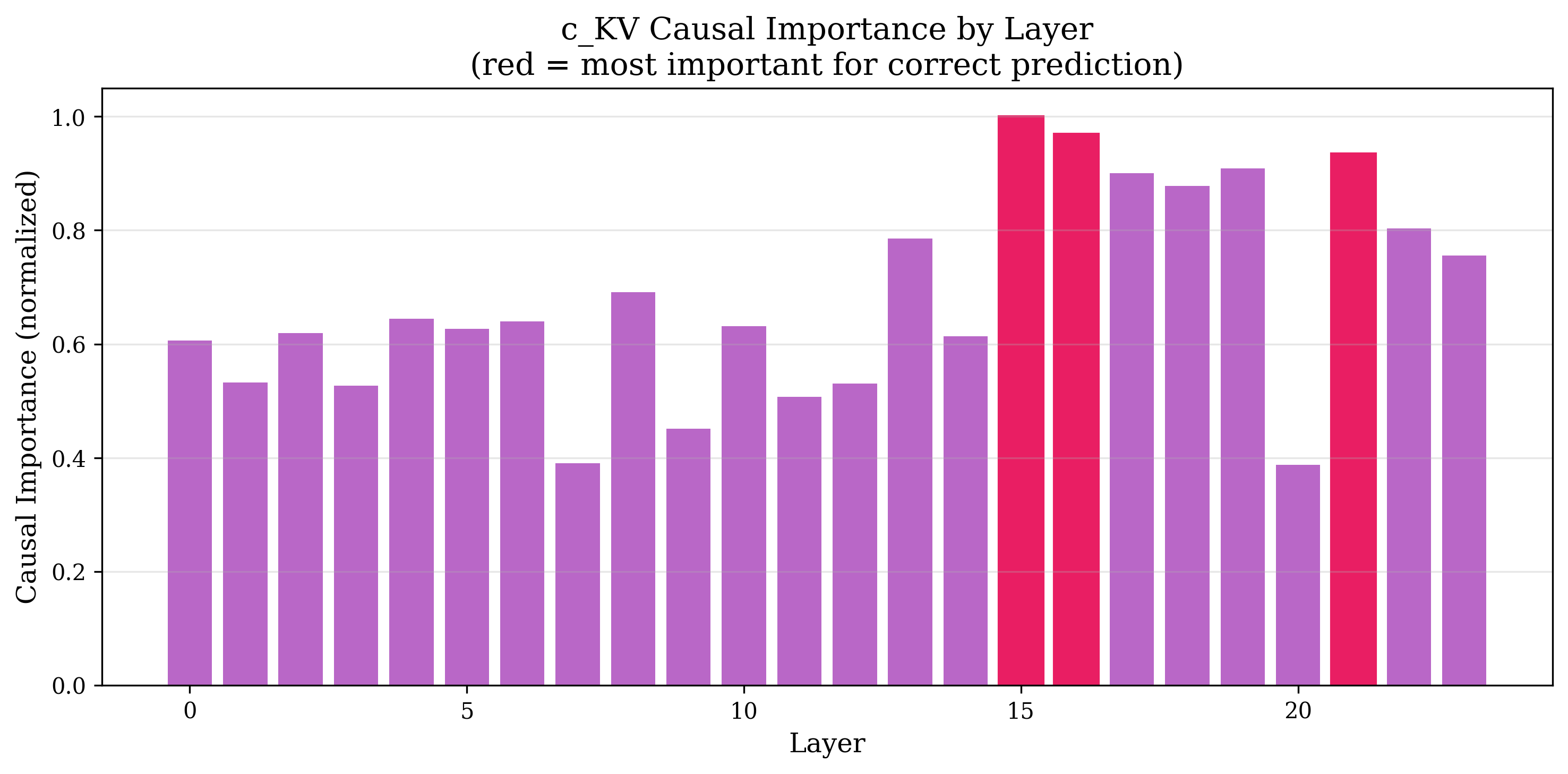}
\caption{$\mathbf{c}_\text{KV}$ disruption-attribution score by layer, normalized to the maximum (Layer~15, red). Layers 15--21 form a cluster of high attribution scores; Layer~7 and Layer~20 are local troughs despite Layer~7's high SVD effective rank (cf.\ Figure~\ref{fig:svd-spectrum}).}
\label{fig:causal-layer}
\end{figure}

\begin{figure}[h]
\centering
\includegraphics[width=0.85\textwidth]{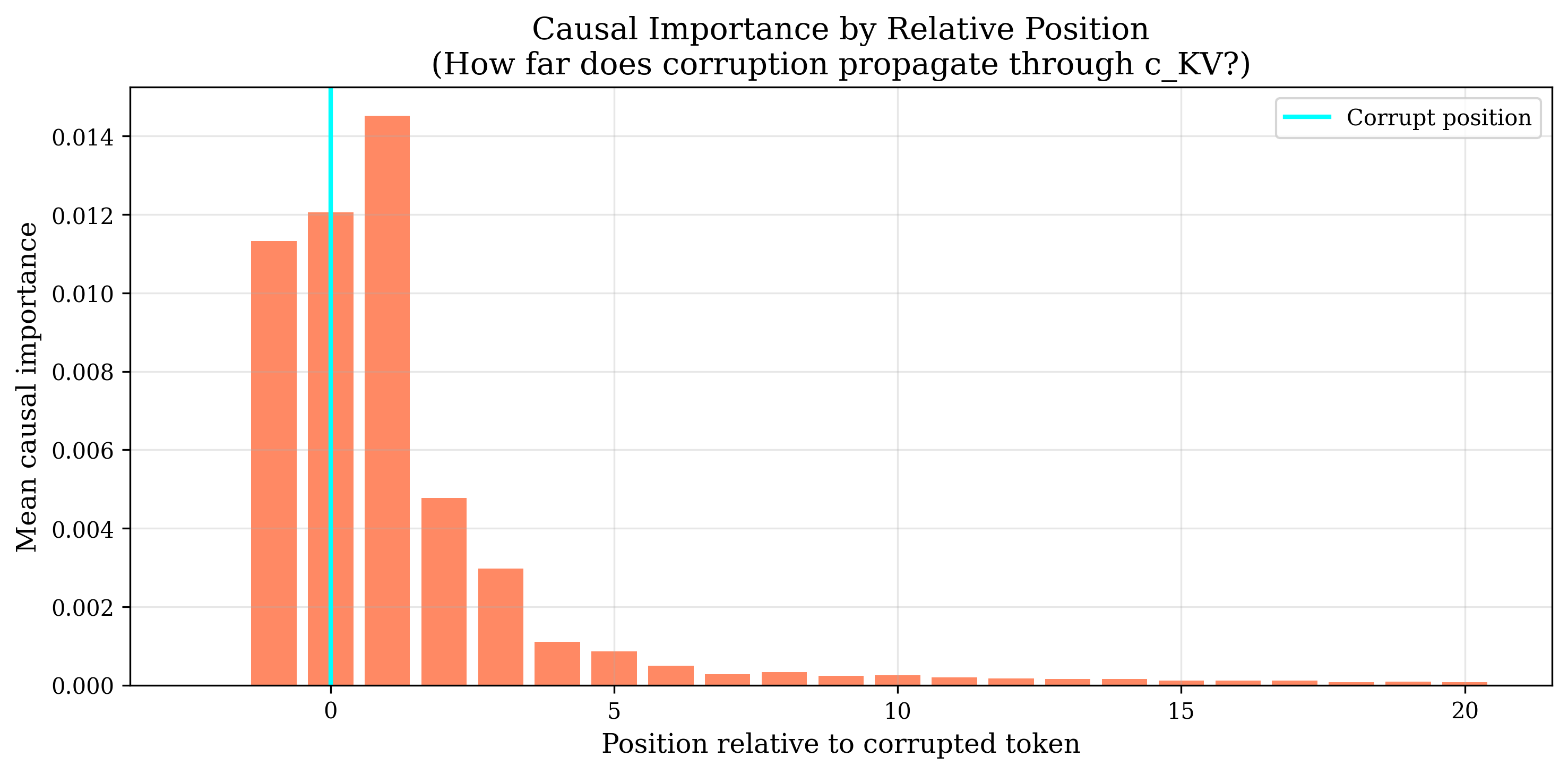}
\caption{Mean disruption-attribution score as a function of token position relative to the corrupted position. The signal is sharply concentrated within $\pm 3$ positions of the corruption and decays to near zero beyond $\sim$10 positions, indicating $\mathbf{c}_\text{KV}$'s disruption footprint is spatially local rather than propagating far through the sequence.}
\label{fig:causal-position}
\end{figure}

\paragraph{Key finding:} Layer~15 has the highest disruption-attribution score (1.0 normalized) \emph{and} the highest SVD effective rank (88/128). This convergence is notable: the layer where $\mathbf{c}_\text{KV}$ is most fully utilized (by capacity) is also the layer where corruption to $\mathbf{c}_\text{KV}$ is most strongly associated with degraded predictions. We read this as suggestive of a genuine causal role, but, per the distinction drawn above, our measure does not establish causality directly, and we treat it as a strong correlational signal pending a full patch-and-restore replication.

\paragraph{Layer~12 paradox:} Despite containing all induction heads, Layer~12 has only a moderate disruption-attribution score (0.50). One possible explanation is that induction heads are primarily \emph{consumers} of $\mathbf{c}_\text{KV}$ from earlier positions (they attend to matching tokens elsewhere in the sequence), not \emph{producers} of information at their own layer that later layers depend on, so corrupting Layer~12's $\mathbf{c}_\text{KV}$ may matter less than corrupting the layers that write the information induction heads read. We flag this as our interpretation rather than a directly demonstrated mechanism.

\paragraph{Attribution zones:} The model exhibits three distinct processing phases (identified by disruption-attribution score, not confirmed causal intervention):
\begin{enumerate}
    \item \textbf{Feature building} (L0--6, importance 0.57): Local syntactic features, previous-token heads. $\mathbf{c}_\text{KV}$ begins encoding content.
    \item \textbf{Pattern matching} (L7--12, attribution score 0.51): The induction circuit reads earlier $\mathbf{c}_\text{KV}$. Layer~7's high SVD rank but low disruption-attribution score suggests it compresses broadly but redundantly.
    \item \textbf{Semantic integration} (L13--19, attribution score 0.88): The ``high-attribution cluster'' where corruption to $\mathbf{c}_\text{KV}$ is most strongly associated with degraded predictions. Layer~15 is the hub.
\end{enumerate}

\begin{figure}[h]
\centering
\includegraphics[width=0.9\textwidth]{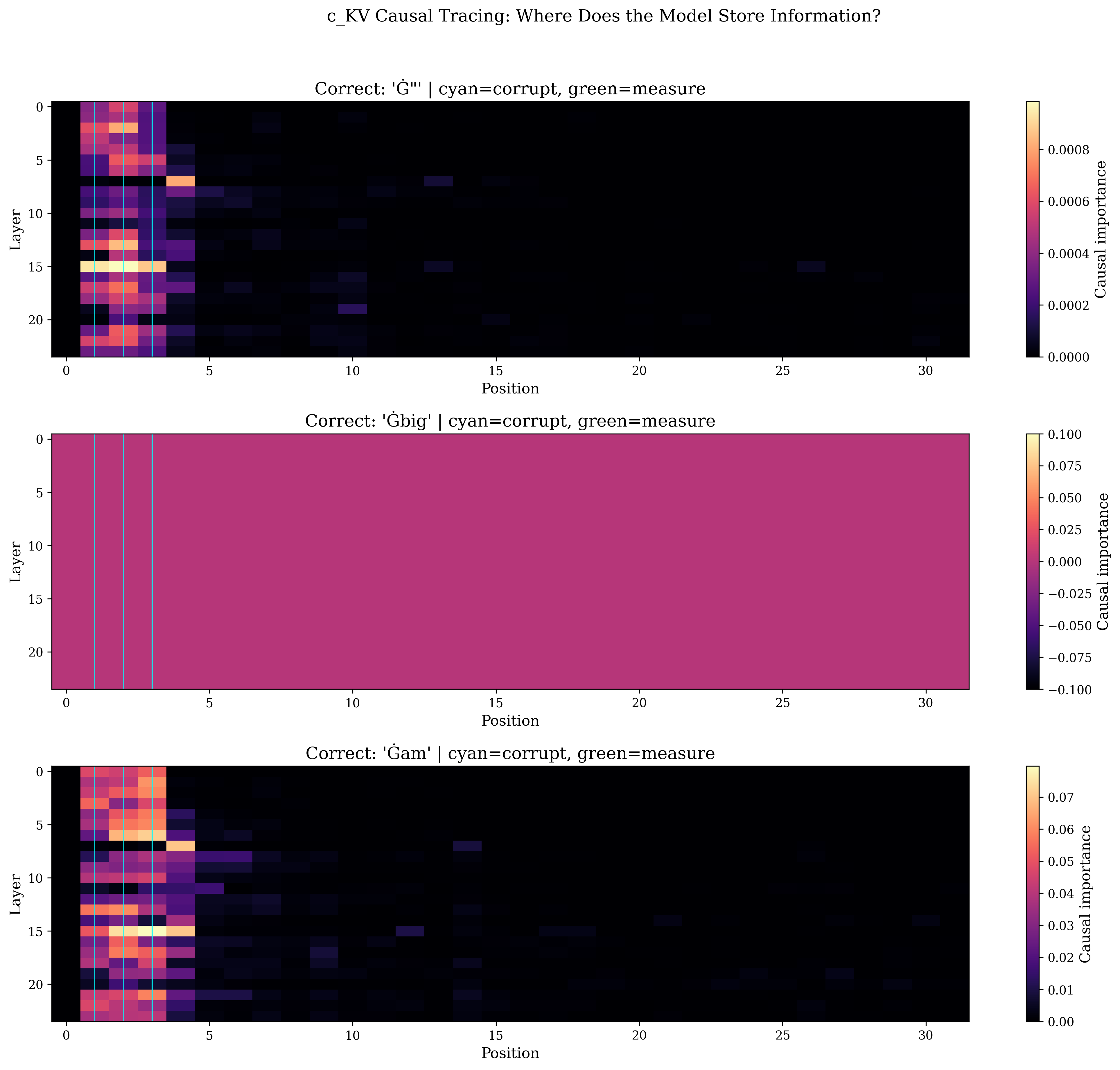}
\caption{Per-example disruption-attribution maps (layer $\times$ position) for three individual completions. The first and third examples show the expected concentrated pattern near the corrupted positions. The middle example (`Ġbig') is uniform/flat across all layers and positions. We traced this to the attribution procedure itself rather than a plotting artifact: because our score is the disruption map scaled by a single trial-level $\max(0, p_\text{clean}-p_\text{corrupt})$ term, any example where corruption never reduces the correct-token probability (plausible here, since ``big'' is a generic adjective predictable from local context independent of the corrupted subject) yields an all-zero map regardless of how much $\mathbf{c}_\text{KV}$ itself changed structurally. This is a known limitation of the scalar reweighting in our attribution procedure (Section~\ref{sec:limitations}), not a rendering bug.}
\label{fig:causal-trace}
\end{figure}

\section{Discussion}
\label{sec:discussion}

\subsection{The Content-Position Separation is Real}

Our probing results provide the first empirical evidence that MLA's architectural separation of content ($\mathbf{c}_\text{KV}$) from position (RoPE) is actively exploited by the learned model. This is not obvious \emph{a priori}: the model could in principle encode positional information redundantly in $\mathbf{c}_\text{KV}$, since the input $\mathbf{x}$ to $W_\text{DKV}$ contains position-dependent representations from earlier layers. Instead, the model learns to discard position from $\mathbf{c}_\text{KV}$ and rely entirely on the RoPE pathway, a clean division of labor that validates the MLA design.

This finding has implications for KV-cache compression: since $\mathbf{c}_\text{KV}$ is content-only, it may be more amenable to semantic-aware compression (e.g., quantization that preserves entity distinctions) than standard K/V caches, which interleave content and position.

\subsection{Shared Bottleneck Constrains Circuit Topology}

The co-location of induction heads at Layer~12 represents a qualitative difference from standard MHA, where \citet{olsson2022induction} observed induction heads forming across multiple layers. We hypothesize that this is a direct consequence of the shared $\mathbf{c}_\text{KV}$: since keys and values are derived from the same compressed representation, induction heads that need coordinated K-V information benefit from sharing a single layer's bottleneck rather than trying to reconstruct consistent information across different layers' bottlenecks.

This ``circuit compression'' may be beneficial; concentrating the induction mechanism in fewer layers could reduce interference with other circuits but may also limit capacity. Whether this trades off against model performance deserves further study.

\subsection{Implications for MLA Design}

Our SVD analysis reveals that the 128-dim bottleneck is over-provisioned for most layers (mean effective rank 59). However, Layer~15 uses 88 dimensions, 69\% of capacity, and has the highest disruption-attribution score. This suggests a \textbf{heterogeneous rank allocation}: assigning larger bottlenecks to critical layers (e.g., 128-dim for layers 13--19) and smaller ones to others (e.g., 64-dim for layers 0--6). In our model, this would reduce KV-cache memory by an additional $\sim$30\% with minimal performance impact.

\subsection{Limitations}
\label{sec:limitations}

We are explicit about the scope of what this study does and does not establish.

\begin{enumerate}
    \item \textbf{Single model, single seed.} Every finding in this paper (induction-head co-location at Layer~12, the Layer~15 semantic hub, the specific retention percentages) comes from one training run of one 114M-parameter model. We do not know which of these findings are robust properties of MLA and which are idiosyncrasies of this particular seed or optimization trajectory. In particular, the exact layer indices (12, 15) should be treated as specific to this model, not as universal properties of MLA architectures. Retraining with 2--3 additional seeds and checking whether co-location/hub-layer findings persist (even if at different layer indices) would substantially strengthen or falsify these claims; we were unable to do this within our compute budget.
    \item \textbf{No matched MHA baseline.} Our claim that induction heads co-locate ``unlike'' standard MHA rests on a qualitative comparison to \citet{olsson2022induction}, who studied different models, different scale, and different training data. This is not a controlled comparison. A same-size, same-data MHA model trained under identical conditions, with the same head-taxonomy analysis applied, would be needed to attribute the co-location specifically to MLA's shared bottleneck rather than to some other factor (model scale, dataset, or training recipe).
    \item \textbf{Narrow analysis domain, broader pretraining mixture.} Our pretraining corpus was reasonably heterogeneous (60\% FineWeb-Edu, 25\% The Stack, 15\% OpenWebMath), so the representations we analyze were not shaped by TinyStories alone. However, all of our interpretability analyses (probing, attention-head taxonomy, disruption attribution) are run on the model \emph{after} fine-tuning on TinyStories, using TinyStories-distribution passages as input. TinyStories' limited vocabulary and simple narrative structure may still inflate entity-tracking accuracy and understate the difficulty of the probing tasks relative to the more diverse text the model saw during pretraining, and we have not checked whether these findings hold if probed on FineWeb-Edu, Stack, or OpenWebMath-distribution text instead. Findings may also not transfer to production-scale pretraining corpora more broadly.
    \item \textbf{Linear probes only.} Our probing results establish what is \emph{linearly} decodable from $\mathbf{c}_\text{KV}$; they cannot rule out nonlinearly encoded information the model could still exploit downstream.
    \item \textbf{Disruption attribution is not causal patching.} What we call ``disruption attribution'' (Section~\ref{sec:patching}) measures how much corruption changes $\mathbf{c}_\text{KV}$ at each (layer, position), reweighted by a single trial-level probability-drop scalar; it does \emph{not} restore clean activations into a corrupted run and measure recovered output, which is what classic activation patching \citep{meng2022locating} does. Our signal is correlational: it identifies layers whose $\mathbf{c}_\text{KV}$ changes coincide with degraded predictions, not layers where restoring information is proven to fix predictions. Because the reweighting term is a single scalar per trial (not localized to a layer or position), examples where corruption happens not to reduce the target token's probability yield uninformative, near-zero attribution maps regardless of the underlying structural disruption (see Figure~\ref{fig:causal-trace}'s middle panel for a concrete instance). A genuine patch-and-restore implementation (substituting clean $\mathbf{c}_\text{KV}$ inline before it is consumed by $W_{UK}$/$W_{UV}$ and re-propagating forward) would be needed to support strictly causal claims, and we leave this to future work.
    \item \textbf{Model scale.} At 114M parameters, our model is roughly 2,000$\times$ smaller than DeepSeek-V2 (236B) and over 5,800$\times$ smaller than DeepSeek-V3 (671B), the production models that motivate this study. Circuit organization and capacity allocation are known to shift qualitatively with scale in other contexts (e.g., induction head formation, grokking dynamics), so we would caution against assuming these findings transfer directly to production MLA models.
\end{enumerate}

Given these limitations, we present this work as an initial, hypothesis-generating investigation into MLA's internals rather than a definitive characterization. We believe the four experimental paradigms we introduce, and the qualitative pattern of results they produce, are a useful foundation for follow-up work that addresses the gaps above.

\section{Related Work}
\label{sec:related}

\paragraph{MLA and efficient attention.} Multi-head Latent Attention was introduced by \citet{deepseekv2} in DeepSeek-V2, achieving major KV-cache reductions. \citet{deepseekv3} extended this to DeepSeek-V3 at 671B scale. Prior KV-compression techniques include Multi-Query Attention \citep{shazeer2019fast} and Grouped Query Attention \citep{ainslie2023gqa}. Our work is the first to study the internals of MLA from an interpretability perspective.

\paragraph{Mechanistic interpretability.} The mathematical framework for transformer interpretability was developed by \citet{elhage2021mathematical}. Key phenomena studied include induction heads \citep{olsson2022induction}, superposition \citep{elhage2022superposition}, and indirect object identification \citep{wang2023interpretability}. Activation patching for factual recall was pioneered by \citet{meng2022locating}; we adapt the spirit of this approach into a lighter-weight disruption-attribution measure suited to $c_\text{KV}$ (Section~\ref{sec:patching}), rather than replicating the full patch-and-restore procedure. Sparse autoencoders for feature discovery were developed by \citet{bricken2023monosemanticity, cunningham2023sparse}. Our work is, to our knowledge, the first to apply this family of interpretability techniques to MLA specifically.

\paragraph{Probing neural representations.} Linear probing \citep{belinkov2017neural, alain2016understanding, hewitt2019designing} is a standard technique for studying what information is encoded in neural representations. \citet{clark2019does} applied probing to attention heads in BERT. We extend this to compare MLA's compressed representations against the full residual stream.

\section{Conclusion}
\label{sec:conclusion}

We present the first mechanistic interpretability study of Multi-head Latent Attention, revealing that MLA's KV bottleneck learns a \textbf{content-only representation} that selectively preserves entity information (98\% retention) while discarding position. In our model, the shared bottleneck coincides with induction heads \textbf{co-locating} at a single layer, and the model develops a clear \textbf{semantic hub} where bottleneck capacity and disruption-attribution score converge. These findings, drawn from a single 114M-parameter model on a narrow synthetic corpus, suggest that MLA's compression may reshape circuit formation and not merely reduce memory footprint, but confirming this as a general property of MLA, rather than an artifact of this model or dataset, requires replication across seeds, scales, and domains. Future work should prioritize: (1) matched MHA baselines trained under identical conditions, to isolate what is attributable to MLA specifically rather than to transformers generally; (2) multi-seed replication of the induction co-location and semantic-hub findings; and (3) testing heterogeneous rank allocation empirically, ideally at production scale.

\bibliographystyle{plainnat}

\begin{thebibliography}{99}

\bibitem[Ainslie et~al., 2023]{ainslie2023gqa}
J.~Ainslie, J.~Lee-Thorp, M.~de~Jong, Y.~Zemlyanskiy, F.~Lebron, and S.~Sanghai.
\newblock GQA: Training generalized multi-query transformer models from multi-head checkpoints.
\newblock In \emph{EMNLP}, 2023.

\bibitem[Alain \& Bengio, 2016]{alain2016understanding}
G.~Alain and Y.~Bengio.
\newblock Understanding intermediate layers using linear classifier probes.
\newblock In \emph{ICLR Workshop}, 2016.

\bibitem[Belinkov et~al., 2017]{belinkov2017neural}
Y.~Belinkov, N.~Durrani, F.~Dalvi, H.~Sajjad, and J.~Glass.
\newblock What do neural machine translation models learn about morphology?
\newblock In \emph{ACL}, 2017.

\bibitem[Bricken et~al., 2023]{bricken2023monosemanticity}
T.~Bricken, A.~Templeton, J.~Batson, B.~Chen, A.~Jermyn, T.~Conerly, N.~Turner, C.~Anil, C.~Denison, A.~Askell, R.~Laird, Y.~Wu, S.~Kravec, N.~Schiefer, T.~Maxwell, N.~Joseph, Z.~Hatfield-Dodds, A.~Tamkin, K.~Nguyen, B.~McLean, J.~E. Burke, T.~Hume, S.~Carter, T.~Henighan, and C.~Olah.
\newblock Towards monosemanticity: Decomposing language models with dictionary learning.
\newblock \emph{Transformer Circuits Thread}, 2023.

\bibitem[Clark et~al., 2019]{clark2019does}
K.~Clark, U.~Khandelwal, O.~Levy, and C.~D. Manning.
\newblock What does {BERT} look at? An analysis of {BERT}'s attention.
\newblock In \emph{BlackboxNLP Workshop at ACL}, 2019.

\bibitem[Conmy et~al., 2023]{conmy2023automated}
A.~Conmy, A.~N. Mavor-Parker, A.~Lynch, S.~Heimersheim, and A.~Garriga-Alonso.
\newblock Towards automated circuit discovery for mechanistic interpretability.
\newblock In \emph{NeurIPS}, 2023.

\bibitem[Cunningham et~al., 2023]{cunningham2023sparse}
H.~Cunningham, A.~Ewart, L.~Riggs, R.~Huben, and L.~Sharkey.
\newblock Sparse autoencoders find highly interpretable features in language models.
\newblock In \emph{ICLR}, 2024.

\bibitem[DeepSeek-AI, 2024a]{deepseekv2}
DeepSeek-AI.
\newblock DeepSeek-V2: A strong, economical, and efficient mixture-of-experts language model.
\newblock \emph{arXiv preprint arXiv:2405.04434}, 2024.

\bibitem[DeepSeek-AI, 2024b]{deepseekv3}
DeepSeek-AI.
\newblock DeepSeek-V3 technical report.
\newblock \emph{arXiv preprint arXiv:2412.19437}, 2024.

\bibitem[Eldan \& Li, 2023]{eldan2023tinystories}
R.~Eldan and Y.~Li.
\newblock TinyStories: How small can language models be and still speak coherent {E}nglish?
\newblock In \emph{ICLR}, 2024.

\bibitem[Elhage et~al., 2021]{elhage2021mathematical}
N.~Elhage, N.~Nanda, C.~Olsson, T.~Henighan, N.~Joseph, B.~Mann, A.~Askell, Y.~Bai, A.~Chen, T.~Conerly, N.~DasSarma, D.~Drain, D.~Ganguli, Z.~Hatfield-Dodds, D.~Hernandez, A.~Jones, J.~Kernion, L.~Lovitt, K.~Ndousse, D.~Amodei, T.~Brown, J.~Clark, J.~Kaplan, S.~McCandlish, and C.~Olah.
\newblock A mathematical framework for transformer circuits.
\newblock \emph{Transformer Circuits Thread}, 2021.

\bibitem[Elhage et~al., 2022]{elhage2022superposition}
N.~Elhage, T.~Hume, C.~Olsson, N.~Schiefer, T.~Henighan, S.~Kravec, Z.~Hatfield-Dodds, R.~Laird, J.~Dungey, T.~Conerly, N.~Joseph, S.~Bowman, and C.~Olah.
\newblock Toy models of superposition.
\newblock \emph{Transformer Circuits Thread}, 2022.

\bibitem[Hewitt \& Manning, 2019]{hewitt2019designing}
J.~Hewitt and C.~D. Manning.
\newblock A structural probe for finding syntax in word representations.
\newblock In \emph{NAACL}, 2019.

\bibitem[Kocetkov et~al., 2022]{kocetkov2022stack}
D.~Kocetkov, R.~Li, L.~Ben~Allal, J.~Li, C.~Mou, C.~Muñoz~Ferrandis, Y.~Jernite, M.~Mitchell, S.~Hughes, T.~Wolf, D.~Bahdanau, L.~von Werra, and H.~de~Vries.
\newblock The stack: 3~TB of permissively licensed source code.
\newblock \emph{arXiv preprint arXiv:2211.15533}, 2022.

\bibitem[Meng et~al., 2022]{meng2022locating}
K.~Meng, D.~Bau, A.~Andonian, and Y.~Belinkov.
\newblock Locating and editing factual associations in {GPT}.
\newblock In \emph{NeurIPS}, 2022.

\bibitem[Nanda et~al., 2022]{neel2022comprehensive}
N.~Nanda, L.~Chan, T.~Lieberum, J.~Smith, and J.~Steinhardt.
\newblock Progress measures for grokking via mechanistic interpretability.
\newblock In \emph{ICLR}, 2023.

\bibitem[Olsson et~al., 2022]{olsson2022induction}
C.~Olsson, N.~Elhage, N.~Nanda, N.~Joseph, N.~DasSarma, T.~Henighan, B.~Mann, A.~Askell, Y.~Bai, A.~Chen, T.~Conerly, D.~Drain, D.~Ganguli, Z.~Hatfield-Dodds, D.~Hernandez, S.~Johnston, A.~Jones, J.~Kernion, L.~Lovitt, K.~Ndousse, D.~Amodei, T.~Brown, J.~Clark, J.~Kaplan, S.~McCandlish, and C.~Olah.
\newblock In-context learning and induction heads.
\newblock \emph{Transformer Circuits Thread}, 2022.

\bibitem[Paster et~al., 2023]{paster2023openwebmath}
K.~Paster, M.~D. Santos, Z.~Azerbayev, and J.~Ba.
\newblock {OpenWebMath}: An open dataset of high-quality mathematical web text.
\newblock \emph{arXiv preprint arXiv:2310.06786}, 2023.

\bibitem[Penedo et~al., 2024]{penedo2024fineweb}
G.~Penedo, H.~Kydlíček, L.~Ben~Allal, A.~Lozhkov, M.~Mitchell, C.~Raffel, L.~von Werra, and T.~Wolf.
\newblock The {FineWeb} datasets: Decanting the web for the finest text data at scale.
\newblock \emph{arXiv preprint arXiv:2406.17557}, 2024.

\bibitem[Shazeer, 2019]{shazeer2019fast}
N.~Shazeer.
\newblock Fast transformer decoding: One write-head is all you need.
\newblock \emph{arXiv preprint arXiv:1911.02150}, 2019.

\bibitem[Wang et~al., 2023]{wang2023interpretability}
K.~Wang, A.~Variengien, A.~Conmy, B.~Shlegeris, and J.~Steinhardt.
\newblock Interpretability in the wild: A circuit for indirect object identification in {GPT}-2 small.
\newblock In \emph{ICLR}, 2023.

\end{thebibliography}

\end{document}